\documentclass[10pt,twocolumn,letterpaper]{article}

\usepackage[pagenumbers]{cvpr}      %

\usepackage[T1]{fontenc}
\usepackage{placeins}

\usepackage{microtype}

\usepackage{xspace}
\newcommand{\method}{SonoAtlasWorld\xspace}
\newcommand{\vol}{\mathcal{V}}            %
\newcommand{\labels}{\mathcal{L}}         %
\newcommand{\pose}{p}                     %
\newcommand{\act}{a}                      %
\newcommand{\mask}{m}                     %
\newcommand{\Seat}{\operatorname{Seat}}
\newcommand{\Render}{\operatorname{R}}
\newcommand{\Exp}{\operatorname{Exp}}
\newcommand{\SE}{\mathrm{SE}(3)}
\newcommand{\E}{\mathbb{E}}
\newcommand{\B}[1]{\mathbf{#1}}           %

\makeatletter
\patchcmd{\paragraph}
  {3.25ex \@plus1ex \@minus.2ex}
  {1.3ex \@plus.3ex \@minus.2ex}
  {}{\PackageWarning{SonoAtlasLayout}{Paragraph spacing patch not applied}}
\makeatother

\definecolor{cvprblue}{rgb}{0.21,0.49,0.74}
\usepackage[breaklinks,colorlinks,allcolors=cvprblue]{hyperref}

\makeatletter
\patchcmd{\@maketitle}{\vskip .375in}{\vskip 0pt}{}{}
\patchcmd{\@maketitle}{\vspace*{24pt}}{\vspace*{4pt}}{}{}
\patchcmd{\@maketitle}{\vspace*{12pt}}{\vspace*{0pt}}{}{}
\makeatother

\title{SonoAtlasWorld: Anatomy-Grounded Ultrasound World Models\\
for Wide-Range Scanning}

\author{
\href{https://openreview.net/profile?id=~Zheyuan_Zhang1}{\textcolor{black}{Zheyuan Zhang}}\quad \href{https://openreview.net/profile?id=~Boris_Mailhe1}{\textcolor{black}{Boris Mailhe}}\quad \href{https://openreview.net/profile?id=~Hang_Yu16}{\textcolor{black}{Hang Yu}}\quad \href{https://openreview.net/profile?id=~Yikang_Liu1}{\textcolor{black}{Yikang Liu}}\\
\href{https://openreview.net/profile?id=~Xiao_Chen12}{\textcolor{black}{Xiao Chen}}\quad \href{https://openreview.net/profile?id=~Ankush_Mukherjee1}{\textcolor{black}{Ankush Mukherjee}}\quad \href{https://openreview.net/profile?id=~Terrence_Chen4}{\textcolor{black}{Terrence Chen}}\quad \href{https://openreview.net/profile?id=~Shanhui_Sun2}{\textcolor{black}{Shanhui Sun}}\thanks{Corresponding author: \href{mailto:shanhui.sun@uii-ai.com}{\texttt{shanhui.sun@uii-ai.com}}.}\\
\href{https://www.uii-ai.com/}{\textcolor{black}{United Imaging Intelligence, Boston, MA}}
}
\date{}

\hypersetup{
  pdftitle={SonoAtlasWorld: Anatomy-Grounded Ultrasound World Models for Wide-Range Scanning},
  pdfauthor={Zheyuan Zhang, Boris Mailhe, Hang Yu, Yikang Liu, Xiao Chen, Ankush Mukherjee, Terrence Chen, Shanhui Sun},
  pdfsubject={Ultrasound world models},
  pdfkeywords={ultrasound, world models, anatomy, robotic ultrasound}
}

\begin{document}
\raggedbottom
\setcounter{dbltopnumber}{1}
\makeatletter
\g@addto@macro\@maketitle{\begin{minipage}{\textwidth}
  \centering
  \includegraphics[width=0.9\linewidth]{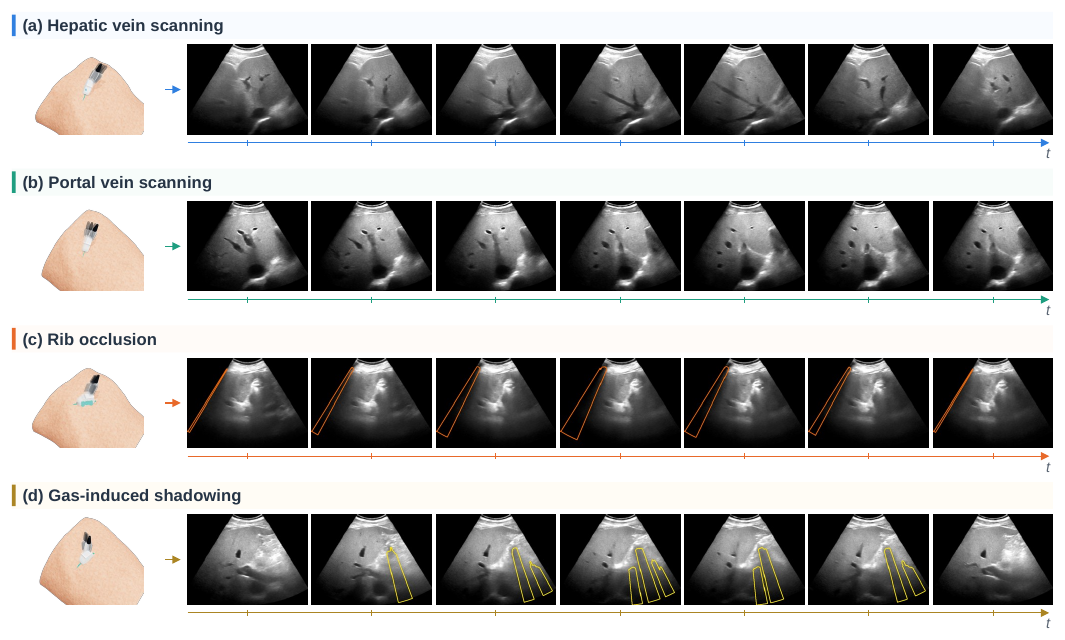}
  \captionsetup{type=figure,skip=4pt}
  \caption{\textbf{From probe motion to evolving ultrasound views.}
  SonoAtlasWorld supports probe-controlled exploration of persistent
  3D anatomy. Each row pairs rendered probe positions
  (left) with generated ultrasound videos by SonoAtlasWorld along the same scan
  (right). Colored outlines indicate shadow regions from the anatomical
  visibility conditions. All ultrasound sequences are generated.}
  \label{fig:teaser}
\end{minipage}
\par\vspace{0pt}
}
\makeatother
\maketitle
\begin{abstract}
Learning robotic ultrasound policies requires interactive environments
for scalable data generation and repeatable evaluation, while exploration
on real patients is limited by acquisition costs and safety constraints.
Generative world models offer an alternative, but image history and probe
actions alone do not ensure anatomical consistency across wide-range
scans and changing acoustic windows. We introduce
\textbf{SonoAtlasWorld}, an ultrasound world model built on
\emph{anatomy as state}. A persistent 3D anatomical asset extracted from
computed tomography (CT) anchors each scan to a specific anatomy.
Probe actions determine anatomical and acoustic visibility conditions
for a causal ultrasound generator trained on clinical videos.
Our three-stage pipeline combines anatomy-conditioned pretraining,
causal conversion and long-rollout alignment. Inference optimizations
achieve pose-to-ultrasound throughput of 81.2 FPS. To evaluate anatomical control
beyond image similarity, we introduce the \textbf{Action-Following Index
(AFI)}, which measures normalized reconstruction gain over a
static-prefix baseline.
On 40 held-out 3D volumes, SonoAtlasWorld achieves liver and macro-organ
AFI of 0.428 and 0.262, compared with at most 0.117 and 0.038 for the
tested action- and geometry-conditioned methods. Long-rollout
alignment reduces brightness drift by $3.46\times$ at the end of 40
repeated scan cycles. SonoAtlasWorld also generates offline policy-training
demonstrations and supports anatomically grounded closed-loop evaluation.
Scaling the generated training data improves key-plane acquisition
success from 48.2\% to 81.2\% on unseen anatomies in simulation,
demonstrating its utility for robotic ultrasound policy learning.
Interactive video demonstrations are available at
\url{https://sonoatlasworld.github.io/SonoAtlasWorld}.
\end{abstract}

\vspace{-3pt}
\section{Introduction}
\label{sec:intro}

Ultrasound provides real-time imaging for diagnosis and intervention,
yet acquiring diagnostic views depends strongly on operator expertise.
Robotic ultrasound offers a route to more reproducible examinations,
reduced physical workload and improved access where experienced
sonographers are scarce~\cite{jiang2023roboticultrasound}. Autonomous
scanning requires policies that coordinate image interpretation and probe
motion across patient anatomies. Their development depends on realistic
interactive environments for generating demonstrations and evaluating
alternative actions.

Constructing these environments remains difficult. Patient scans provide
clinical observations, but repeated trial-and-error exploration is
constrained by safety and practical considerations. Phantoms offer
controlled interaction with limited anatomical diversity; physics-based
simulators must balance image fidelity and computational
cost~\cite{burger2013gpu,salehi2015simulation,ao2025sonogym}. Re-slicing reconstructed
ultrasound volumes remains bounded by the acquired anatomy and viewing
angles~\cite{lasso2014plus}. Learning from recorded scans also faces a
data bottleneck: routine videos lack probe poses, and synchronized
tracking requires calibration and dedicated acquisition
protocols~\cite{yang2026sonoworld,yang2026ultraworld}. Generative world models offer a data-driven approach to interactive
simulation, as illustrated by DreamDojo and
Cosmos~\cite{gao2026dreamdojo,nvidia2025cosmos}. In ultrasound, SonoWorld
grounds probe actions through EM-tracked clinical
scans~\cite{yang2026sonoworld}, whereas UltraWorld uses synthetic
supervision and geometric conditioning without real pose
annotations~\cite{yang2026ultraworld}. 

A central challenge is preserving a specific anatomy throughout
interaction, including when observation history provides an incomplete or
ambiguous spatial reference. In wide-range liver scanning, a large organ
and branching vasculature are viewed through changing cross-sections and
acoustic windows. A vessel may leave the imaging plane or disappear behind
rib or gas shadowing, then reappear from another acoustic window. Recent
history may therefore contain only partial or acoustically obscured views,
while distant anatomical regions can have similar appearances. Models relying solely
on image history and probe actions must infer both unobserved anatomy and
its correspondence across these views. Such cases challenge the preservation of anatomical identity
through learned action--appearance correlations alone, motivating a
persistent 3D anatomical reference during interaction.

\begin{figure*}[t]
  \centering
  \includegraphics[width=\textwidth]{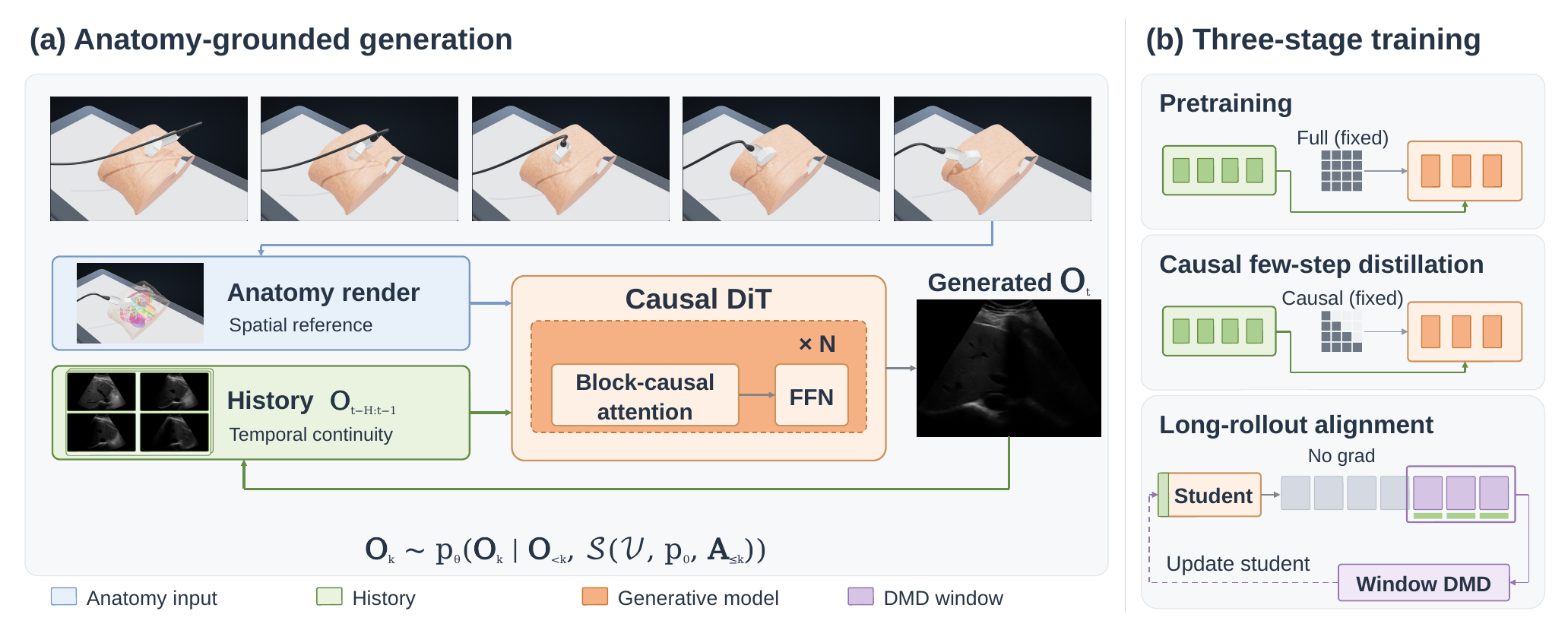}
  \caption{\textbf{Overview of SonoAtlasWorld.} (a) Five rendered poses illustrate interaction with one persistent scene. The final pose, anatomical rendering and generated observation share the same saved time step. The current pose provides anatomy and visibility conditions; generated history provides temporal context for causal ultrasound generation. Persistent anatomy supports consistent modeling under challenging bone-induced occlusion. (b) Training proceeds from anatomy-conditioned pretraining to causal few-step distillation and long-rollout alignment. The final stage applies DMD to a teacher-length window after a no-gradient generated prefix.}
  \label{fig:overview}
  \vspace{-4pt}
\end{figure*}

We introduce \textbf{SonoAtlasWorld}, an anatomy-grounded ultrasound world
model built on the principle of \emph{anatomy as state}. A 3D anatomical
asset persists throughout the scan, and probe actions determine the
cross-section and acoustic visibility sampled from it. A learned causal
observation model generates ultrasound from these conditions and its
observation history. Clinical videos provide appearance and temporal
priors, while the retained anatomical asset provides the spatial reference
for new scan trajectories. Anatomy and visibility conditioning support
exploration across vessels, ribs and gas shadows
(\cref{fig:teaser}). To realize this anatomy-grounded formulation as an interactive environment,
we develop a three-stage learning framework that combines an
anatomy-conditioned ultrasound prior, a causal observation generator and
long-rollout alignment (\cref{fig:overview}). The resulting model supports anatomically consistent, interactive
ultrasound generation with reduced long-rollout drift.

Our contributions are summarized as follows:
\begin{itemize}
    \item We introduce \emph{anatomy as persistent world state}, separating
    persistent patient-specific spatial state from framewise observation
    conditioning. Probe actions dynamically sample spatially consistent
    anatomy and acoustic visibility from this shared 3D state across scan
    trajectories.
    \item We develop a three-stage training pipeline for causal generation
    and long-rollout alignment, and accelerate pose-to-ultrasound
    generation to 81.2 FPS for real-time wide range interactive scanning.
    \item We introduce an anatomy-referenced evaluation protocol for
    ultrasound world models, assessing anatomical action following,
    cross-trajectory consistency, and long-horizon stability. We further
    introduce the Action-Following Index (AFI) to measure anatomical
    agreement beyond a static-prefix baseline, complementing conventional
    image-quality metrics.
    \item We validate SonoAtlasWorld as a source of offline policy-training
    data and a closed-loop evaluation environment, demonstrating policy
    generalization across held-out patient anatomies in simulation.
\end{itemize}

\vspace{-3pt}
\section{Related Work}
\label{sec:related_work}

\paragraph{Generative World Models.}
Genie and AdaWorld learn latent actions from videos for interactive generation
and adaptable control~\cite{bruce2024genie,gao2025adaworld}.
Video foundations such as Wan provide generative
priors~\cite{wan2025}. For temporal generation, teacher forcing uses
ground-truth histories~\cite{williams1989teacherforcing}, Diffusion Forcing
trains with independent per-token noise levels~\cite{chen2024diffusionforcing},
CausVid distills a bidirectional teacher into a few-step causal
student~\cite{yin2025causvid}, Self Forcing trains on generated
histories~\cite{huang2025selfforcing}, and Self-Forcing++ supervises sampled
windows of longer rollouts~\cite{cui2025selfforcingpp}.
Complementing these advances in temporal generation, LiveWorld maintains a
3D background and entities that evolve out of view~\cite{duan2026liveworld},
while PERSIST uses a latent 3D environment, camera state and renderer for
persistent spatial memory~\cite{garcin2026persist}.
SonoAtlasWorld grounds ultrasound observations in a fixed anatomical volume
and probe-dependent visibility, combining a persistent spatial reference
with causal temporal generation.

\paragraph{Medical World Models.}
Medical world modeling spans observation synthesis and representation
learning. Endora and Surgical Vision World Model generate endoscopic and
latent-action-controlled surgical videos,
respectively~\cite{li2024endora,koju2025surgicalwm}, while
semantic-map-conditioned diffusion synthesizes echocardiography
videos~\cite{phi2024echosynthesis}. For representation learning, CheXWorld
uses JEPA for anatomy- and domain-aware radiograph
embeddings~\cite{yue2025chexworld}; Cardiac Copilot, Structure-aware World
Model and EchoWorld learn spatial and motion representations for probe
guidance~\cite{jiang2024cardiaccopilot,jiang2024structure,yue2025echoworld}.
SonoGraph-WM predicts anatomical scene graphs for
planning~\cite{li2026sonographwm}. For interactive ultrasound generation,
SonoWorld grounds clinical-video dynamics with EM-tracked
motion~\cite{yang2026sonoworld}, whereas UltraWorld transfers
anatomy-guided synthetic supervision to a sampling-geometry-conditioned
predictor~\cite{yang2026ultraworld}. Building on controllable medical
synthesis~\cite{zhang2025diffboost,mao2025medsegfactory}, SonoAtlasWorld
retains anatomy and visibility conditions to anchor causal generation in
a shared 3D state.

\paragraph{Interactive Ultrasound Environments.}
Phantoms, physics-based simulation and tracked volume reconstruction
support controlled ultrasound acquisition and robotic
experimentation~\cite{salehi2015simulation,lasso2014plus}.
NVIDIA's ray-based ultrasound simulator renders anatomical geometry but
omits diffraction, interference and
phase~\cite{nvidia2026ultrasoundsimulator}.
Learned image translation improves the appearance of anatomical
simulations~\cite{vitale2020simulation}.
SonoGym combines 3D patient anatomy with parallel robot learning,
physics-based simulation and learned slice-to-ultrasound
rendering~\cite{ao2025sonogym}. Its learned renderer uses pix2pix~\cite{isola2017pix2pix} trained on
paired ex-vivo spine data, whereas our observation model learns temporal
generation from clinical videos under anatomy and visibility conditions.

\vspace{-3pt}
\section{Method}
\label{sec:method}
\begingroup
\setlength{\abovedisplayskip}{4pt plus 1pt minus 1pt}
\setlength{\belowdisplayskip}{4pt plus 1pt minus 1pt}
\setlength{\abovedisplayshortskip}{3pt plus 1pt minus 1pt}
\setlength{\belowdisplayshortskip}{3pt plus 1pt minus 1pt}

\method samples anatomy and visibility from a persistent 3D state under
probe actions and causally generates ultrasound conditioned on these
observation history.

\vspace{-3pt}
\subsection{Data}
\label{sec:data}
We collect 82{,}158 abdominal B-mode clips from 1{,}254 patients
under IRB approval. The patient-disjoint split comprises 78{,}530 training
clips from 1{,}191 patients and 3{,}628 test clips from 63 patients.
After quality filtering, we use 493 unique anatomical assets from public 3D
volumes in AbdomenCT-1K~\cite{ma2022abdomenct1k}, AbdomenAtlas~\cite{li2024abdomenatlas},
and Merlin~\cite{blankemeier2026merlin}, split into 453 training volumes and
40 held-out test volumes.
Clinical videos provide appearance supervision, while 3D assets support
anatomical sampling and policy experiments. Dataset composition,
preprocessing and splits are detailed in Supplementary Sec.~A.1.

\vspace{-3pt}
\subsection{Anatomy as State}
\label{sec:formulation}
\label{sec:simulator}
We represent patient-specific anatomy as a 3D label volume $\vol$ that
remains fixed throughout a scan. At step $t$, a probe-frame action
$\act_t\in\mathbb{R}^6$ specifies a relative translation and rotation,
updating the probe pose $\pose_t\in\SE$ and its sampled condition:
\begin{align}
  \pose_t &= \Seat_{\vol}\bigl(\pose_{t-1}\,\Exp(\act_t)\bigr),
  \label{eq:transition}\\
  \mask_t &= \Render(\vol,\pose_t),\quad
  \mask_t\in\labels^{H\times W}.
  \label{eq:render}
\end{align}
The operator $\Seat_{\vol}$ restores probe--skin contact, while $\Render$
samples anatomy and acoustic visibility within the imaging fan using the
shared organ, vessel and visibility vocabulary $\labels$.
For causal generation, we group ultrasound frames and conditions into blocks $\B{O}_k$ and
$\B{M}_k$. Given an initial pose $\pose_0$ and probe actions
$\B{A}_{\leq k}$ through block $k$, the sampler $\mathcal{S}$ composes
\cref{eq:transition,eq:render} to produce the condition sequence
$\B{M}_{\leq k}$. The causal observation model is
\begin{equation}
  \B{O}_k \sim p_\theta\bigl(\B{O}_k\mid
  \B{O}_{<k},\mathcal{S}(\vol,\pose_0,\B{A}_{\leq k})\bigr).
  \label{eq:wm}
\end{equation}
The anatomical volume $\vol$ defines the persistent patient-specific spatial
state, while the sampler maps probe actions to pose-dependent anatomical
conditions $\B{M}_{\leq k}$ supplied to the generator without direct
processing of the full 3D volume.
This separation preserves anatomical consistency across arbitrary probe
trajectories and revisited acoustic windows through the state and sampler,
while the generated observation history provides temporal continuity.

To construct $\vol$, we combine abdominal organ, vessel, rib and
body-surface annotations with automated
segmentation~\cite{wasserthal2023totalsegmentator} and additional
intrahepatic vessel models~\cite{isensee2021nnunet,xu2025vsnet,antonelli2022msd}. At each updated pose, $\Render$ extracts a
curvilinear cross-section and traces beam-wise obstruction. Bone, costal
cartilage and contact loss produce \emph{acoustic dropout}; intraluminal
gas produces a distinct \emph{gas shadow} condition. These labels
distinguish intersected anatomy from acoustically hidden structures.
Real and rendered annotations share $\labels$,
allowing the generator to learn from clinical videos and accept sampled
3D conditions at inference. Asset preprocessing and visibility rules are
detailed in Supplementary Sec.~A.3.

\vspace{-3pt}
\subsection{Three-Stage World-Model Training}
\label{sec:wm}
We train the observation model in three stages:

\paragraph{Pretraining: ultrasound prior and anatomy conditioning.}
\label{sec:bidirectional}
We first adapt a latent video diffusion transformer~\cite{wan2025} to
clinical ultrasound with a fixed text prompt. We then add an
anatomy-conditioning branch initialized from the first $L_c$ backbone
blocks. The branch encodes mask videos with the shared video autoencoder
and injects residuals into the backbone through zero-initialized
projections~\cite{zhang2023controlnet}:
\begin{equation}
  \B{h}_\ell \leftarrow \B{h}_\ell
  + Z_\ell\bigl(\B{c}_\ell(\B{M})\bigr),
  \label{eq:branch}
\end{equation}
where $\ell=1,\ldots,L_c$, $\B{c}_\ell$ is the condition feature, and
$Z_\ell$ is its projection. The branch has no noise-timestep input.
For a clean latent $\B{x}_0$ and noise
$\boldsymbol\epsilon\sim\mathcal{N}(0,I)$, let
$\B{x}_s=(1-s)\B{x}_0+s\boldsymbol\epsilon$.
We use the flow-matching objective~\cite{lipman2023flow,liu2023rectified}:
\begin{equation}
  \mathcal{L}_{\mathrm{FM}}
  = \E\Bigl[
      \lambda(s)\,
      \bigl\|v_\theta(\B{x}_s,s,\B{M})
      -(\boldsymbol\epsilon-\B{x}_0)\bigr\|_{w}^{2}
    \Bigr].
  \label{eq:fm}
\end{equation}
Here, $s$ is the noise level, $\lambda(s)$ is the timestep weight, and
$\|\cdot\|_w^2$ is a normalized squared error with spatial weights
$w=w(\B{M})$ that emphasize small organs. Backbone and branch are optimized
jointly. Paired fan augmentation transforms each clinical clip and its
annotations together to accommodate different acquisition geometries.

\paragraph{Middle training: causal conversion and few-step distillation.}
We replace bidirectional attention with block-causal attention: each
block attends to its history and all tokens within the current block.
The condition branch operates per block to prevent future-mask leakage.
We first fine-tune with flow matching using teacher-forced context.
Following causal video distillation~\cite{yin2025causvid,huang2025selfforcing},
we initialize a four-step student from teacher-generated trajectories.
For a saved teacher state $\B{x}_{s_i}^{\mathrm{T}}$ and its final clean sample
$\B{x}_0^{\mathrm{T}}$, the student's clean prediction $d_\theta$ minimizes
\begin{equation}
  \mathcal{L}_{\mathrm{init}}
  = \E\Bigl[
      \bigl\|d_\theta(\B{x}_{s_i}^{\mathrm{T}},s_i,\B{M})
      -\B{x}_0^{\mathrm{T}}\bigr\|_2^2
    \Bigr],
  \label{eq:init}
\end{equation}
with $i$ sampled from four noise levels. This initializes few-step
sampling without requiring the student to reproduce the full teacher
trajectory.

We then train on \emph{self-generated history} using distribution matching
distillation (DMD)~\cite{yin2024dmd,huang2025selfforcing}. Let
$\hat{\B{x}}_0=G_\theta(\B{M})$ be an autoregressive student rollout,
and $\hat{\B{x}}_s$ its re-noised version. A frozen bidirectional teacher
$f_{\mathrm{T}}$ and a trainable critic $f_\psi$ predict clean latents from
the shared input $\B{z}_s=(\hat{\B{x}}_s,s,\B{M})$. With the student
Jacobian $\B{J}_\theta=\partial\hat{\B{x}}_0/\partial\theta$, DMD gives
\begin{equation}
  \nabla_\theta\mathcal{L}_{\mathrm{DMD}}
  = \E\biggl[
      \B{J}_\theta^{\top}\,
      \frac{f_\psi(\B{z}_s)-f_{\mathrm{T}}(\B{z}_s)}{c_s}
    \biggr],
  \label{eq:dmd}
\end{equation}
where $c_s$ normalizes the update magnitude. Teacher and critic
predictions are detached for this update; the critic is trained separately
to denoise generated clips. This
stage uses anatomy sequences without paired ultrasound targets, exposing
the student to the history it encounters during autoregressive generation.

\begin{table*}[!t]
  \centering
  \caption{\textbf{Patient-specific anatomical fidelity, action following and temporal continuity.} Given anatomical inputs, SonoAtlasWorld's benefits extend beyond gross liver structure to finer anatomical recovery and temporal continuity. Vascular distance: median 2D error from projected reference branch points to the predicted vessel skeleton, conditional on detection within 10\,mm. Warping errors: 8-bit gray levels; prefix-only (--) omitted for nearly static output. Bold/italic: best/second-best excluding prefix-only and teacher. \emph{Real}/\emph{sim}: real/generated training data. $^\dagger$Adapted re-implementations; real action/geometry baselines use EM tracking.}
  \label{tab:anatomy_e1}
  \fontencoding{T1}\selectfont
  \small
  \setlength{\tabcolsep}{4pt}
  \begin{tabular}{@{}llcccccc@{\hspace{14pt}}ccc@{}}
    \toprule
    Method & Data & \multicolumn{6}{c}{General Scanning} & \multicolumn{3}{c}{Rib Occlusion} \\
    \cmidrule(lr){3-8}\cmidrule(lr){9-11}
    & & \multicolumn{2}{c}{Liver} & \multicolumn{2}{c}{Macro-organ} & Vascular & Temporal & Shadow & Recovery & Temporal \\
    \cmidrule(lr){3-4}\cmidrule(lr){5-6}
    & & Dice $\uparrow$ & AFI $\uparrow$ & Dice $\uparrow$ & AFI $\uparrow$ & Dist.\ (mm) $\downarrow$ & $E_{\mathrm{warp}}\downarrow$ & Contrast $\uparrow$ & Liver Dice $\uparrow$ & $E_{\mathrm{warp}}\downarrow$ \\
    \midrule
    \multicolumn{11}{@{}l}{\emph{No control}} \\
    Prefix only & -- & 0.670 & 0 & 0.219 & 0 & 3.62 & -- & 0.132 & 0.645 & -- \\
    \midrule
    \multicolumn{11}{@{}l}{\emph{Action- or geometry-conditioned world models}} \\
    SonoWorld~\cite{yang2026sonoworld}$^\dagger$ & real & 0.662 & 0.107 & 0.220 & 0.032 & 3.33 & \textit{2.39} & 0.106 & 0.681 & 3.06 \\
    SonoWorld~\cite{yang2026sonoworld}$^\dagger$ & sim & 0.605 & 0.084 & 0.208 & 0.031 & 3.66 & 2.50 & 0.085 & 0.666 & 2.82 \\
    UltraWorld~\cite{yang2026ultraworld}$^\dagger$ & real & 0.667 & 0.117 & 0.224 & 0.038 & 3.40 & 2.52 & 0.123 & 0.669 & 3.26 \\
    UltraWorld~\cite{yang2026ultraworld}$^\dagger$ & sim & 0.518 & 0.041 & 0.171 & 0.020 & 3.42 & 2.44 & 0.073 & 0.633 & \textit{2.62} \\
    \midrule
    \multicolumn{11}{@{}l}{\emph{Frame-conditioned rendering}} \\
    SonoGym~\cite{ao2025sonogym}$^\dagger$ & real & 0.811 & \textit{0.434} & 0.301 & 0.200 & 2.93 & 4.91 & \textit{0.735} & \textbf{0.931} & 6.20 \\
    \midrule
    \multicolumn{11}{@{}l}{\emph{Teacher model}} \\
    \emph{Bidirectional teacher} & real & 0.838 & 0.472 & 0.352 & 0.250 & 2.10 & 2.58 & 0.788 & 0.890 & 2.74 \\
    \midrule
    \multicolumn{11}{@{}l}{\emph{Causal world model}} \\
    Framewise variants & real & \textbf{0.815} & \textbf{0.443} & \textit{0.352} & \textit{0.258} & \textit{1.83} & 2.86 & 0.646 & \textit{0.924} & 4.02 \\
    \textbf{SonoAtlasWorld} & \textbf{real} & \textit{0.813} & 0.428 & \textbf{0.356} & \textbf{0.262} & \textbf{1.75} & \textbf{1.90} & \textbf{0.773} & 0.911 & \textbf{2.47} \\
    \bottomrule
  \end{tabular}
  \vspace{-6pt}
\end{table*}

\paragraph{Post-training: long-rollout alignment.}
\label{sec:method_long}
Short distillation clips provide limited exposure to cache eviction and
accumulated generation errors. We therefore continue training with the
rolling inference cache~\cite{cui2025selfforcingpp} over horizons longer than the teacher's. Let
$H_{\mathrm{T}}$ be the teacher horizon in blocks. We draw a nominal rollout
length $N$ and a block-aligned window start $j$ using discrete uniform
sampling $\mathcal{U}_{\mathrm{d}}$:
\begin{equation}
  N \sim \mathcal{U}_{\mathrm{d}}(N_{\min},N_{\max}),\qquad
  j \sim \mathcal{U}_{\mathrm{d}}(0,N-H_{\mathrm{T}}).
  \label{eq:long_window}
\end{equation}
Here, $\mathcal{U}_{\mathrm{d}}(a,b)$ is uniform over integers $a,\ldots,b$.
We apply \cref{eq:dmd} to blocks $j+1,\ldots,j+H_{\mathrm{T}}$ with their
aligned anatomy conditions. Earlier blocks are
generated without gradients. Causality allows generation to stop at the
window endpoint, avoiding unnecessary later blocks. The window's first
frame and mask sequence are re-encoded as a fresh clip to match the
teacher's temporal autoencoder layout. This extends the contexts used for
training while keeping teacher supervision within its original horizon.

\paragraph{Streaming generation.}
At inference, the key--value cache retains an initial attention
sink~\cite{xiao2024sink} and a fixed-length recent history; older states
are evicted and rotary positions compacted~\cite{su2024rope}. Causal
autoencoder caches carry features across blocks. Each block requires four
denoising steps followed by a clean-cache commit pass, enabling continued
generation as probe commands arrive. The main model uses post-trained EMA
weights. Architecture, noise schedules and cache settings are provided in
Supplementary Sec.~A.
\endgroup

\begin{figure*}[t]
  \centering
  \vspace{-6pt}
  \includegraphics[width=0.82\textwidth]{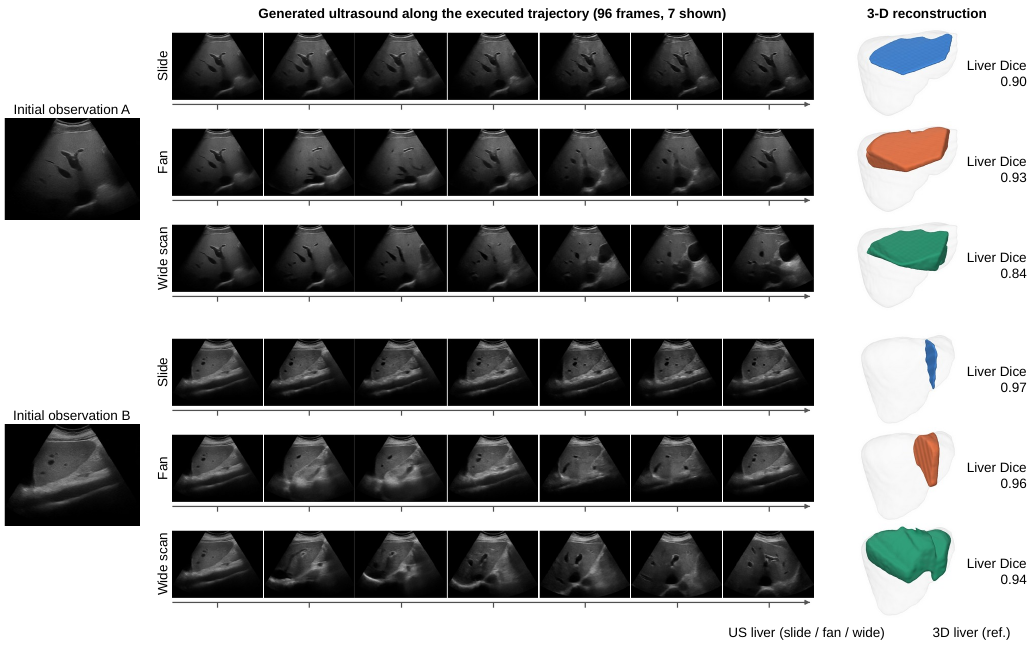}
  \caption{\textbf{Anatomical consistency across controlled scans.}
  For each of two initial observations, slide, fan and wide-scan
  trajectories produce distinct ultrasound sequences. Right: liver
  reconstructions from generated frames along the executed poses,
  overlaid on the reference 3D anatomy. The shared anatomical state
  grounds each scan in the same spatial reference.}
  \label{fig:e1_branch}
  \vspace{-10pt}
\end{figure*}

\vspace{-3pt}
\section{Experiments}
\label{sec:experiments}

\vspace{-3pt}
\subsection{Anatomical Action Following}
\label{sec:exp_anatomy}
Image metrics such as PSNR, SSIM and
LPIPS~\cite{zhang2018lpips} are widely used to evaluate
ultrasound world models~\cite{yang2026sonoworld,yang2026ultraworld}, but
image similarity alone does not establish whether generated anatomy
follows the executed probe actions. We therefore reconstruct segmented
generated anatomy along the executed poses and measure its Dice overlap
with the reference 3D anatomy within the visible swept volume.
The segmentation network is trained on independent real ultrasound images.
Anatomical recovery from generated frames also provides an indirect check
of compatibility with real ultrasound appearance.
To discount agreement already explained by the initial observation, we introduce the
Action-Following Index (AFI):
\vspace{-3pt}
\begin{equation}
  \mathrm{AFI}
  = \max\left\{
      0,\;
      \frac{D - D_{\mathrm{prefix}}}
           {D_{\mathrm{ref}} - D_{\mathrm{prefix}}}
    \right\},
  \label{eq:afi}
\end{equation}
\vspace{-3pt}
where $D_{\mathrm{prefix}}$ is reconstruction Dice when the prefix image
is repeated along the same trajectory, and $D_{\mathrm{ref}}$ is the
anatomy-mask-stream reference. AFI measures the fraction of reconstruction
gain beyond this static baseline recovered under the executed actions.
Macro-organ scores average over the liver, intrahepatic portal vein, hepatic
veins, inferior vena cava, gallbladder and right
kidney, subject to visibility and coverage criteria detailed in the
supplementary material.
Temporal continuity is assessed using optical-flow warping
error $E_{\mathrm{warp}}$~\cite{liu2024evalcrafter}, without paired real video.

All models follow the same trajectories on 40 held-out 3D volumes.
SonoWorld uses relative probe actions and UltraWorld sampling geometry,
trained on limited real EM-tracked data (\emph{real}) or generated
trajectories (\emph{sim}). Given neither receives
patient-specific anatomy at inference, these comparisons validate the
benefit of retaining explicit anatomy rather than generator improvements
under matched inputs. SonoGym renders anatomical labels framewise;
our framewise variant uses SonoAtlasWorld's DiT weights and anatomy inputs
without generated history. The bidirectional teacher is reported separately
to distinguish anatomical fidelity from causal streaming.

\begin{figure*}[t]
  \centering
  \vspace{-6pt}
  \includegraphics[width=0.7\textwidth]{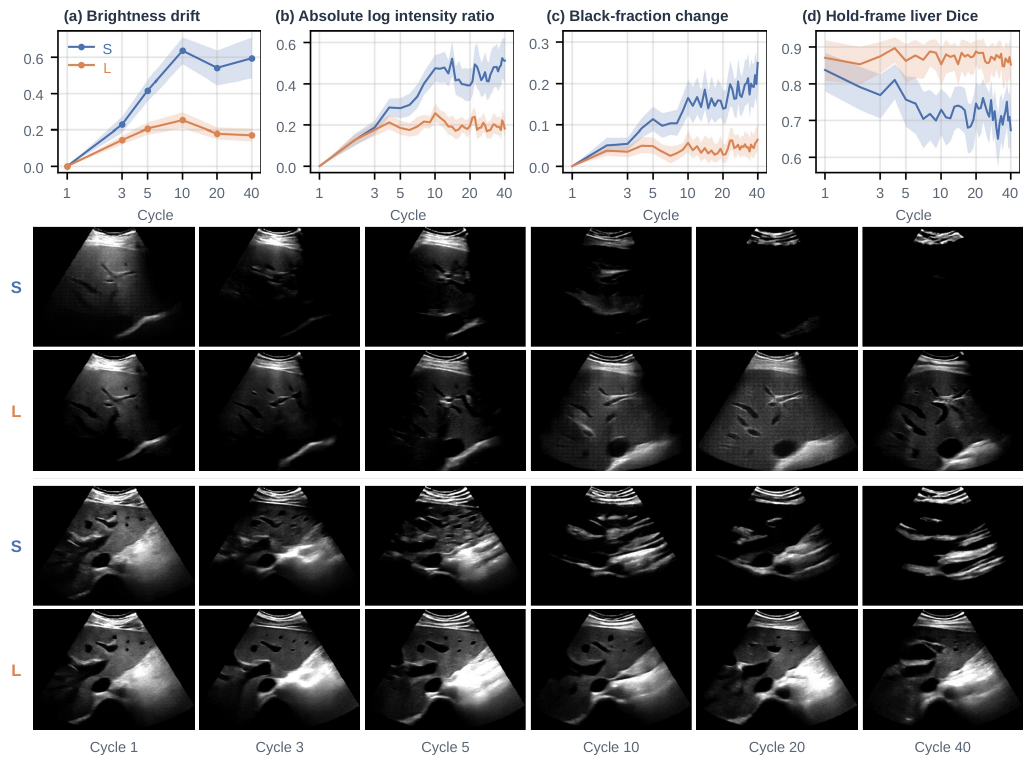}
  \caption{\textbf{Long-rollout alignment mitigates accumulated drift.}
  Compared with the short-rollout model (S), the aligned model (L) better
  preserves ultrasound appearance and anatomy over 40 repeated scan cycles,
  reducing cycle-40 brightness drift by $3.46\times$.
  Top: drift and anatomical agreement across cycles; shaded bands denote
  95\% confidence intervals. Bottom: generated frames at the same return
  pose for two cases. Early appearances are similar, but S progressively
  darkens and loses visible structures, while L maintains clearer boundaries.}
  \label{fig:long_nosink}
  \vspace{-10pt}
\end{figure*}

\paragraph{Anatomical action following.}
Action- and geometry-conditioned models reach at most 0.117 liver AFI and
0.038 macro-organ AFI, with liver Dice below the prefix-only baseline
(\cref{tab:anatomy_e1}). Both the bidirectional teacher and SonoAtlasWorld
significantly outperform these world-model methods on all four anatomical metrics. Thus, probe
actions and sampling geometry alone do not guarantee agreement with the
underlying anatomy, even with simulated training data.
A likely contributor to the limited action following is that clinical scans
typically follow standardized acquisition patterns, restricting action
coverage and generalization to wide-range probe motions.
SonoAtlasWorld achieves the highest macro-organ Dice and AFI
(0.356/0.262), compared with 0.352/0.258 for its framewise variant and
0.301/0.200 for SonoGym. 
SonoAtlasWorld attains a median projected vessel distance of 1.75\,mm
among detected points, providing complementary evidence of patient-specific
vascular alignment (Supplementary Sec.~B.1). The bidirectional
teacher achieves higher liver scores, while SonoAtlasWorld combines
comparable macro-organ agreement with causal, interactive generation.
\Cref{fig:e1_branch} illustrates anatomy recovered along distinct scans.

\paragraph{Cross-trajectory consistency.}
To assess cross-trajectory anatomical consistency, we independently generate
slide, fan and wide-scan trajectories for each of 40 held-out anatomies
and reconstruct their observations in a common CT coordinate system.
We measure pairwise 3D Dice within jointly observed regions and
reconstruction Dice against the reference CT.
SonoAtlasWorld achieves liver cross-trajectory/CT Dice of 0.836/0.813,
comparable to the bidirectional teacher (0.834/0.838) and our framewise
variant (0.820/0.815), with higher CT agreement than action-conditioned
baselines.
Copying the initial frame yields higher cross-trajectory Dice (0.895)
but lower CT Dice (0.659), demonstrating that self-consistency alone
does not establish anatomical correctness.
Together, these results support spatial agreement across independently
generated trajectories and correspondence with the shared patient-specific
anatomy.

\paragraph{Visibility transitions and temporal continuity.}
We further test anatomical consistency as the probe moves from visible
liver into a rib shadow and from the shadow back to visible anatomy
(\cref{tab:anatomy_e1}). Action- and geometry-conditioned models achieve
at most 0.123 shadow contrast on entering occlusion and 0.681 liver Dice
after leaving it, whereas SonoAtlasWorld reaches 0.773 and 0.911,
respectively. SonoGym attains the highest recovery Dice (0.931), but
larger temporal error than our framewise variant (6.20 versus 4.02) and
SonoAtlasWorld (2.47). With identical DiT weights and anatomy inputs,
observation history lowers general-scanning and rib-transition
$E_{\mathrm{warp}}$ from 2.86/4.02 to 1.90/2.47 while preserving liver
Dice (0.813 versus 0.815). This matched comparison supports the role of
causal observation history in improving temporal continuity while
maintaining anatomical fidelity. Protocols are detailed in
Supplementary Sec.~B.1.

\vspace{-3pt}
\subsection{Ultrasound Quality and Condition Alignment}
\label{sec:exp_real}
\begin{table}[!t]
  \centering
  \caption{\textbf{Ultrasound appearance on held-out clinical videos.} Distillation improves perceptual and distributional fidelity. Copy-prefix attains competitive short-clip image scores without generating anatomical change, motivating the complementary AFI evaluation.}
  \label{tab:real_e2a}
  \fontencoding{T1}\selectfont
  \small
  \setlength{\tabcolsep}{3pt}
  \resizebox{\columnwidth}{!}{%
  \begin{tabular}{l c c c c}
    \toprule
    Method & LPIPS $\downarrow$ & PSNR (dB) $\uparrow$ & SSIM $\uparrow$ & DINO-KID-excess $\downarrow$ \\
    \midrule
    Copy-prefix & 0.232 & 21.13 & 0.557 & 0.012 \\
    \midrule
    Causal conversion & 0.333 & 23.45 & 0.633 & 1.403 \\
    Few-step initialization & 0.306 & 21.59 & 0.571 & 0.740 \\
    Short-rollout DMD & 0.210 & 21.33 & 0.602 & 0.177 \\
    \midrule
    SonoAtlasWorld & 0.220 & 21.38 & 0.605 & 0.209 \\
    \bottomrule
  \end{tabular}}
  \vspace{-6pt}
\end{table}

\paragraph{Ultrasound quality on real video.}
We assess ultrasound appearance on held-out clinical clips,
conditioning on a real prefix and estimated anatomy annotations
(\cref{tab:real_e2a}). Few-step initialization and training with generated
history reduce DINO-KID-excess~\cite{binkowski2018kid,simeoni2025dinov3}
from 1.403 to 0.740 and 0.177,
respectively. SonoAtlasWorld achieves LPIPS 0.220, PSNR 21.38\,dB,
SSIM 0.605 and DINO-KID-excess 0.209, supporting reasonable ultrasound
appearance fidelity on real-video continuations. Copy-prefix achieves competitive image scores without anatomical change,
while causal conversion has the highest PSNR/SSIM but the worst LPIPS and
distributional error. These contrasts motivate the complementary AFI
evaluation of anatomical action following (\cref{sec:exp_anatomy});
appearance protocols are detailed in Supplementary Sec.~B.3.

\paragraph{CT-to-US anatomical condition alignment.}
To characterize the domain shift between ultrasound-derived training
conditions and CT-rendered inference conditions, we compare anatomical
geometry at matched left-lobe--aorta (LLA) views, a standard view with
clearly delineated anatomical structures. We report medians across
59 depth-matched US clip--CT volume pairs, using one frame per clip or
volume (Supplementary Sec.~B.4). After matching the aortic
depth, which is used solely as an alignment criterion, liver depth extent
is similar between real-US and CT-derived masks (62.8 vs.\ 64.5\,mm), as
is its normalized depth extent relative to the
skin-to-aorta distance (0.680 vs.\ 0.709). Liver solidity is also close
(0.923 vs.\ 0.940). Residual differences remain in the liver inferior
boundary (2.0\,mm) and centroid depth (4.1\,mm). These results indicate
that CT-rendered masks reproduce several key geometric characteristics
of real ultrasound conditioning at matched views, while not excluding
differences in fine-scale
anatomy.

\vspace{-3pt}

\subsection{Long-Rollout Behaviour}
\label{sec:exp_long}
Stable long rollouts are essential for ultrasound world models to support
extended, interactive wide-range scanning. Prior work evaluates round-trip consistency
and rollouts beyond the training horizon~\cite{yang2026sonoworld}, but
accumulated drift over many repeated revisits remains insufficiently
characterized. We therefore introduce a repeated-cycle evaluation to
quantify this long-rollout drift. The probe follows the same closed path and returns to its
starting pose after each cycle without resetting generation. Comparing
these repeated observations with the first cycle separates temporal drift
from changes caused by probe motion. We measure brightness drift,
absolute log intensity ratio, black-pixel fraction change and liver Dice
at the return pose. Metric definitions and evaluation details are provided
in Supplementary Sec.~B.5.

We compare the short-rollout model (S) with its long-rollout-aligned
variant (L) over 40 cycles (120\,s of generated video at 16\,FPS), using only recent generated history
(\cref{fig:long_nosink}). At cycle 40, L reduces
brightness drift from 0.594 to 0.172, a $3.46\times$ reduction (71\%),
and absolute log intensity change from 0.512 to 0.182 ($2.81\times$).
Liver Dice at the return pose is 0.852 for L versus 0.674 for S.
The qualitative comparison shows similar appearance in the first cycle,
but S progressively darkens and loses visible anatomy, whereas L better
preserves tissue appearance over extended generation. These results
show substantially greater resistance to accumulated drift for L than S
when generation relies on the model's own history.

\vspace{-3pt}
\subsection{Inference Speed Benchmark}
\label{sec:exp_speed}
Although the bidirectional teacher can improve generation quality, it jointly
denoises an entire clip and achieves only 17.6 FPS, limiting its
suitability for interactive environments
(\cref{tab:inference_speed}). We accelerate the causal world model in two
steps: static KV buffers and CUDA graphs reduce generation time from
295.2 to 175.1\,ms/block, and TAEHV decoding~\cite{boerbohan2025taehv} further reduces it to
130.5\,ms/block (92.0 FPS), yielding a $2.26\times$ speedup over the vanilla
causal model. Moving anatomy rendering from CPU to GPU
reduces sampling time from 305.7 to 16.0\,ms. Together, these optimizations
enable pose-to-ultrasound generation at 81.2 FPS, supporting real-time
interactive scanning. For 60\,Hz and 90\,Hz control inputs, the output
throughput is bounded by 60 and 81.2 FPS, respectively, with a
steady-state median pose-to-first-frame latency of 255.0 and 250.7\,ms,
as generation proceeds in 12-frame blocks following the blockwise autoregressive setup of
CausVid and Self Forcing~\cite{yin2025causvid,huang2025selfforcing}.
All timings are measured on a single NVIDIA
H100 GPU and benchmark protocols are provided in the supplementary material.

\begin{table}[!ht]
  \centering
  \caption{\textbf{Inference speed improvements.} Static KV buffers, CUDA graphs and TAEHV decoding accelerate the causal world model by $2.26\times$. GPU anatomy sampling enables pose-to-US generation at 81.2 FPS.}
  \label{tab:inference_speed}
  \small
  \setlength{\tabcolsep}{3pt}
  \begin{tabular}{@{}lrr@{}}
    \toprule
    Configuration & ms/block $\downarrow$ & FPS $\uparrow$ \\
    \midrule
    Bidirectional teacher & 682.8 & 17.6 \\
    \midrule
    \multicolumn{3}{@{}l}{\emph{Causal world model}} \\
    Vanilla (eager, Wan VAE) & 295.2 & 40.6 \\
    + Static KV / CUDA graphs & 175.1 & 68.5 \\
    + TAEHV decoder & 130.5 & \textbf{92.0} \\
    \midrule
    \multicolumn{3}{@{}l}{\emph{Anatomy sampling}} \\
    CPU renderer & 305.7 & -- \\
    GPU renderer & 16.0 & -- \\
    \midrule
    \multicolumn{3}{@{}l}{\emph{End to end}} \\
    \textbf{Pose-to-US} & \textbf{147.7} & \textbf{81.2} \\
    \bottomrule
  \end{tabular}
  \vspace{-6pt}
\end{table}

\begin{figure}[t]
  \centering
  \includegraphics[width=.9\columnwidth]{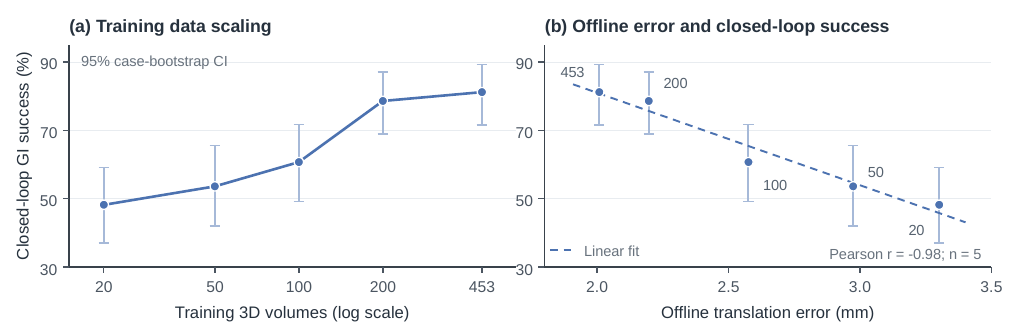}
  \caption{\textbf{World-model data scaling improves downstream policy learning.}
  (a) More training volumes and generated trajectories improve closed-loop
  key-plane acquisition on unseen volumes. (b) Lower offline action-prediction
  error is associated with higher closed-loop success across the displayed
  checkpoints.}
  \label{fig:dp_scaling}
  \vspace{-6pt}
\end{figure}

\vspace{-3pt}
\section{World Model for Policy Learning}
\label{sec:downstream}
Acquiring standard anatomical planes is a central task in ultrasound
scanning. To evaluate \method as a source of offline robotic policy
training data, we sample expert trajectories through 3D volumes targeting
three liver views: the portal-vein long-axis plane, the first hepatic hilum plane and
the second hepatic hilum plane. \method generates the corresponding
ultrasound observations, yielding 39.2k demonstration trajectories across
453 training 3D volumes. We train a goal-conditioned diffusion
policy~\cite{chi2023diffusionpolicy} offline to predict probe actions from
these observations and probe-state information. We then freeze the policy
and evaluate it in the world model through closed-loop interaction,
generating new observations in response to its actions.
Training and evaluation use disjoint 3D
volumes; evaluation on held-out volumes therefore tests generalization
to unseen anatomy. Success is determined from generated ultrasound by a
frozen key-plane classifier trained on a large corpus of real ultrasound
images, which identifies whether the target plane has been acquired.

With the policy architecture and training budget fixed, scaling from
20 to 453 training volumes (1.7k to 39.2k trajectories) raises closed-loop
success from 48.2\% to 81.2\%, reaching 78.6\% at 200 volumes
(\cref{fig:dp_scaling}a). Offline validation translation error decreases
from 3.30 to 2.01\,mm and correlates with success across the displayed
checkpoints ($r=-0.98$; \cref{fig:dp_scaling}b). Together with the
anatomy-following results, these findings support scalable demonstration
generation and controlled policy evaluation on unseen anatomies in
simulation. Online policy training and transfer to physical scanning
remain future work. Full results and protocols are provided in
Supplementary Sec.~D.

\vspace{-3pt}
\section{Conclusion}
\label{sec:conclusion}
SonoAtlasWorld grounds causal ultrasound generation in a persistent 3D
anatomical state and probe-dependent visibility.
Experiments demonstrate anatomical action following, long-rollout
stability, 81.2 FPS generation, and policy learning and evaluation on
unseen anatomies in simulation. However, several limitations remain. Reliance on static anatomical assets
limits the modeling of probe-induced tissue deformation and physiological
motion. The CT-to-US domain gap and segmenter-dependent anatomical
evaluation warrant further investigation. Policy validation remains
confined to simulation; future work will evaluate transfer to robotic
hardware under appropriate safety controls.

\FloatBarrier %
\clearpage
\section*{LLM Usage Statement}
OpenAI GPT and Claude assisted manuscript editing, literature inspection, and code editing. 
The authors remain responsible for all scientific claims, code, and results.


{
    \small
    
    \begin{thebibliography}{51}
\providecommand{\natexlab}[1]{#1}
\providecommand{\url}[1]{\texttt{#1}}
\expandafter\ifx\csname urlstyle\endcsname\relax
  \providecommand{\doi}[1]{doi: #1}\else
  \providecommand{\doi}{doi: \begingroup \urlstyle{rm}\Url}\fi

\bibitem[Antonelli et~al.(2022)Antonelli, Reinke, Bakas,
  et~al.]{antonelli2022msd}
Michela Antonelli, Annika Reinke, Spyridon Bakas, et~al.
\newblock The medical segmentation decathlon.
\newblock \emph{Nature Communications}, 13:\penalty0 4128, 2022.

\bibitem[Ao et~al.(2025)Ao, Moghani, Mittal, Prajapat, Wu, Giraud, Carrillo,
  Krause, and F\"{u}rnstahl]{ao2025sonogym}
Yunke Ao, Masoud Moghani, Mayank Mittal, Manish Prajapat, Luohong Wu, Frederic
  Giraud, Fabio Carrillo, Andreas Krause, and Philipp F\"{u}rnstahl.
\newblock Sonogym: High performance simulation for challenging surgical tasks
  with robotic ultrasound.
\newblock In \emph{Advances in Neural Information Processing Systems}, 2025.

\bibitem[Bi{\'n}kowski et~al.(2018)Bi{\'n}kowski, Sutherland, Arbel, and
  Gretton]{binkowski2018kid}
Miko{\l}aj Bi{\'n}kowski, Danica~J. Sutherland, Michael Arbel, and Arthur
  Gretton.
\newblock Demystifying {MMD GANs}.
\newblock In \emph{International Conference on Learning Representations}, 2018.

\bibitem[Blankemeier et~al.(2026)Blankemeier, Kumar, Cohen, Liu, Liu, Van~Veen,
  Gardezi, Yu, Paschali, Chen, et~al.]{blankemeier2026merlin}
Louis Blankemeier, Ashwin Kumar, Joseph~Paul Cohen, Jiaming Liu, Longchao Liu,
  Dave Van~Veen, Syed Jamal~Safdar Gardezi, Hongkun Yu, Magdalini Paschali,
  Zhihong Chen, et~al.
\newblock {Merlin}: a computed tomography vision--language foundation model and
  dataset.
\newblock \emph{Nature}, 652\penalty0 (8112):\penalty0 1318--1328, 2026.

\bibitem[Boer~Bohan(2025)]{boerbohan2025taehv}
Ollin Boer~Bohan.
\newblock {TAEHV}: Tiny autoencoder for hunyuan video.
\newblock \url{https://github.com/madebyollin/taehv}, 2025.
\newblock Accessed: 2026-11-13.

\bibitem[Bruce et~al.(2024)Bruce, Dennis, Edwards, Parker-Holder, Shi, Hughes,
  Lai, Mavalankar, Steigerwald, Apps, et~al.]{bruce2024genie}
Jake Bruce, Michael Dennis, Ashley Edwards, Jack Parker-Holder, Yuge Shi,
  Edward Hughes, Matthew Lai, Aditi Mavalankar, Richie Steigerwald, Chris Apps,
  et~al.
\newblock Genie: Generative interactive environments.
\newblock In \emph{International Conference on Machine Learning (ICML)}, 2024.

\bibitem[B{\"u}rger et~al.(2013)B{\"u}rger, Bettinghausen, R{\"a}dle, and
  Hesser]{burger2013gpu}
Benny B{\"u}rger, Sascha Bettinghausen, Matthias R{\"a}dle, and J{\"u}rgen
  Hesser.
\newblock Real-time {GPU}-based ultrasound simulation using deformable mesh
  models.
\newblock \emph{IEEE Transactions on Medical Imaging}, 32\penalty0
  (3):\penalty0 609--618, 2013.

\bibitem[Chen et~al.(2024)Chen, Mart{\'i}~Mons{\'o}, Du, Simchowitz, Tedrake,
  and Sitzmann]{chen2024diffusionforcing}
Boyuan Chen, Diego Mart{\'i}~Mons{\'o}, Yilun Du, Max Simchowitz, Russ Tedrake,
  and Vincent Sitzmann.
\newblock {Diffusion Forcing}: Next-token prediction meets full-sequence
  diffusion.
\newblock In \emph{Advances in Neural Information Processing Systems}, 2024.

\bibitem[Chi et~al.(2023)Chi, Feng, Du, Xu, Cousineau, Burchfiel, and
  Song]{chi2023diffusionpolicy}
Cheng Chi, Siyuan Feng, Yilun Du, Zhenjia Xu, Eric Cousineau, Benjamin
  Burchfiel, and Shuran Song.
\newblock Diffusion policy: Visuomotor policy learning via action diffusion.
\newblock In \emph{Robotics: Science and Systems (RSS)}, 2023.

\bibitem[Cui et~al.(2025)Cui, Wu, Li, Yang, Li, Wang, Bai, Ban, and
  Hsieh]{cui2025selfforcingpp}
Justin Cui, Jie Wu, Ming Li, Tao Yang, Xiaojie Li, Rui Wang, Andrew Bai,
  Yuanhao Ban, and Cho-Jui Hsieh.
\newblock Self-forcing++: Towards minute-scale high-quality video generation.
\newblock \emph{arXiv preprint arXiv:2510.02283}, 2025.

\bibitem[Duan et~al.(2026)Duan, Xia, Zhang, Zhang, Zhou, Gou, He, Chen, Zhang,
  and Liu]{duan2026liveworld}
Zicheng Duan, Jiatong Xia, Zeyu Zhang, Wenbo Zhang, Gengze Zhou, Chenhui Gou,
  Yefei He, Feng Chen, Xinyu Zhang, and Lingqiao Liu.
\newblock Liveworld: Simulating out-of-sight dynamics in generative video world
  models.
\newblock \emph{arXiv preprint arXiv:2603.07145}, 2026.

\bibitem[Gao et~al.(2025)Gao, Zhou, Du, Zhang, and Gan]{gao2025adaworld}
Shenyuan Gao, Siyuan Zhou, Yilun Du, Jun Zhang, and Chuang Gan.
\newblock Adaworld: Learning adaptable world models with latent actions.
\newblock In \emph{International Conference on Machine Learning (ICML)}, 2025.

\bibitem[Gao et~al.(2026)Gao, Liang, Zheng, Malik, Ye, Yu, Tseng, Dong, Mo,
  Lin, et~al.]{gao2026dreamdojo}
Shenyuan Gao, William Liang, Kaiyuan Zheng, Ayaan~Naveed Malik, Seonghyeon Ye,
  Sihyun Yu, Wei-Cheng Tseng, Yuzhu Dong, Kaichun Mo, Chen-Hsuan Lin, et~al.
\newblock {DreamDojo}: A generalist robot world model from large-scale human
  videos.
\newblock In \emph{ICML}, pages 33340--33366. PMLR, 2026.

\bibitem[Garcin et~al.(2026)Garcin, Walker, McDonagh, Pearce, Bilen, He, Wang,
  and Bian]{garcin2026persist}
Samuel Garcin, Thomas Walker, Steven McDonagh, Tim Pearce, Hakan Bilen, Tianyu
  He, Kaixin Wang, and Jiang Bian.
\newblock Beyond pixel histories: World models with persistent 3d state.
\newblock In \emph{International Conference on Machine Learning (ICML)}, 2026.

\bibitem[Huang et~al.(2025)Huang, Li, He, Zhou, and
  Shechtman]{huang2025selfforcing}
Xun Huang, Zhengqi Li, Guande He, Mingyuan Zhou, and Eli Shechtman.
\newblock Self forcing: Bridging the train-test gap in autoregressive video
  diffusion.
\newblock In \emph{Advances in Neural Information Processing Systems}, 2025.

\bibitem[Isensee et~al.(2021)Isensee, Jaeger, Kohl, Petersen, and
  Maier-Hein]{isensee2021nnunet}
Fabian Isensee, Paul~F. Jaeger, Simon A.~A. Kohl, Jens Petersen, and Klaus~H.
  Maier-Hein.
\newblock {nnU-Net}: A self-configuring method for deep learning-based
  biomedical image segmentation.
\newblock \emph{Nature Methods}, 18:\penalty0 203--211, 2021.

\bibitem[Isola et~al.(2017)Isola, Zhu, Zhou, and Efros]{isola2017pix2pix}
Phillip Isola, Jun-Yan Zhu, Tinghui Zhou, and Alexei~A. Efros.
\newblock Image-to-image translation with conditional adversarial networks.
\newblock In \emph{CVPR}, pages 1125--1134, 2017.

\bibitem[Jiang et~al.(2024{\natexlab{a}})Jiang, Li, Sun, Jia, Sun, Luo, Song,
  and Huang]{jiang2024structure}
Haojun Jiang, Meng Li, Zhenguo Sun, Ning Jia, Yu Sun, Shaqi Luo, Shiji Song,
  and Gao Huang.
\newblock Structure-aware world model for probe guidance via large-scale
  self-supervised pre-train.
\newblock In \emph{Simplifying Medical Ultrasound (ASMUS), MICCAI Workshop},
  pages 58--67, 2024{\natexlab{a}}.

\bibitem[Jiang et~al.(2024{\natexlab{b}})Jiang, Sun, Jia, Li, Sun, Luo, Song,
  and Huang]{jiang2024cardiaccopilot}
Haojun Jiang, Zhenguo Sun, Ning Jia, Meng Li, Yu Sun, Shaqi Luo, Shiji Song,
  and Gao Huang.
\newblock {Cardiac Copilot}: Automatic probe guidance for echocardiography with
  world model.
\newblock In \emph{Medical Image Computing and Computer Assisted Intervention
  -- MICCAI 2024}, pages 190--199, 2024{\natexlab{b}}.

\bibitem[Jiang et~al.(2023)Jiang, Salcudean, and
  Navab]{jiang2023roboticultrasound}
Zhongliang Jiang, Septimiu~E. Salcudean, and Nassir Navab.
\newblock Robotic ultrasound imaging: State-of-the-art and future perspectives.
\newblock \emph{Medical Image Analysis}, 89:\penalty0 102878, 2023.

\bibitem[Koju et~al.(2025)Koju, Bastola, Shrestha, Amgain, Shrestha, Poudel,
  and Bhattarai]{koju2025surgicalwm}
Saurabh Koju, Saurav Bastola, Prashant Shrestha, Sanskar Amgain, Yash~Raj
  Shrestha, Rudra P.~K. Poudel, and Binod Bhattarai.
\newblock Surgical vision world model.
\newblock In \emph{MICCAI Workshop on Data Engineering in Medical Imaging
  (DEMI)}, 2025.

\bibitem[Lasso et~al.(2014)Lasso, Heffter, Rankin, Pinter, Ungi, and
  Fichtinger]{lasso2014plus}
Andras Lasso, Tamas Heffter, Adam Rankin, Csaba Pinter, Tamas Ungi, and Gabor
  Fichtinger.
\newblock {PLUS}: Open-source toolkit for ultrasound-guided intervention
  systems.
\newblock \emph{IEEE Transactions on Biomedical Engineering}, 61\penalty0
  (10):\penalty0 2527--2537, 2014.

\bibitem[Li et~al.(2024{\natexlab{a}})Li, Liu, Liu, Feng, Li, Liu, Chen, Shao,
  and Yuan]{li2024endora}
Chenxin Li, Hengyu Liu, Yifan Liu, Brandon~Y. Feng, Wuyang Li, Xinyu Liu, Zhen
  Chen, Jing Shao, and Yixuan Yuan.
\newblock Endora: Video generation models as endoscopy simulators.
\newblock In \emph{Medical Image Computing and Computer Assisted Intervention
  (MICCAI)}, pages 230--240, 2024{\natexlab{a}}.

\bibitem[Li et~al.(2024{\natexlab{b}})Li, Qu, Chen, Bassi, Shi, Lai, Yu, Xue,
  Chen, Lin, et~al.]{li2024abdomenatlas}
Wenxuan Li, Chongyu Qu, Xiaoxi Chen, Pedro R. A.~S. Bassi, Yijia Shi, Yuxiang
  Lai, Qian Yu, Huimin Xue, Yixiong Chen, Xiaorui Lin, et~al.
\newblock Abdomenatlas: A large-scale, detailed-annotated, and multi-center
  dataset for efficient transfer learning and open algorithmic benchmarking.
\newblock \emph{Medical Image Analysis}, 2024{\natexlab{b}}.

\bibitem[Li et~al.(2026)Li, Chen, Li, Jiang, Navab, and Bi]{li2026sonographwm}
Xuesong Li, Shuai Chen, Feng Li, Zhongliang Jiang, Nassir Navab, and Yuan Bi.
\newblock Scanning while imagining: A scene-graph world model for robotic
  ultrasound navigation.
\newblock \emph{arXiv preprint arXiv:2609.32837}, 2026.

\bibitem[Lipman et~al.(2023)Lipman, Chen, Ben-Hamu, Nickel, and
  Le]{lipman2023flow}
Yaron Lipman, Ricky T.~Q. Chen, Heli Ben-Hamu, Maximilian Nickel, and Matt Le.
\newblock Flow matching for generative modeling.
\newblock In \emph{ICLR}, 2023.

\bibitem[Liu et~al.(2023)Liu, Gong, and Liu]{liu2023rectified}
Xingchao Liu, Chengyue Gong, and Qiang Liu.
\newblock Flow straight and fast: Learning to generate and transfer data with
  rectified flow.
\newblock In \emph{ICLR}, 2023.

\bibitem[Liu et~al.(2024)Liu, Cun, Liu, Wang, Zhang, Chen, Liu, Zeng, Chan, and
  Shan]{liu2024evalcrafter}
Yaofang Liu, Xiaodong Cun, Xuebo Liu, Xintao Wang, Yong Zhang, Haoxin Chen,
  Yang Liu, Tieyong Zeng, Raymond Chan, and Ying Shan.
\newblock Evalcrafter: Benchmarking and evaluating large video generation
  models.
\newblock In \emph{CVPR}, pages 22139--22149, 2024.

\bibitem[Ma et~al.(2022)Ma, Zhang, Gu, Zhu, Ge, Zhang, An, Wang, Wang, Liu,
  et~al.]{ma2022abdomenct1k}
Jun Ma, Yao Zhang, Song Gu, Cheng Zhu, Cheng Ge, Yichi Zhang, Xingle An,
  Congcong Wang, Qiyuan Wang, Xin Liu, et~al.
\newblock Abdomenct-1k: Is abdominal organ segmentation a solved problem?
\newblock \emph{IEEE Transactions on Pattern Analysis and Machine
  Intelligence}, 44\penalty0 (10):\penalty0 6695--6714, 2022.

\bibitem[Mao et~al.(2025)Mao, Wang, Tang, Xu, Wang, Yang, Zhou, and
  Zhou]{mao2025medsegfactory}
Jiawei Mao, Yuhan Wang, Yucheng Tang, Daguang Xu, Kang Wang, Yang Yang, Zongwei
  Zhou, and Yuyin Zhou.
\newblock {MedSegFactory}: Text-guided generation of medical image-mask pairs.
\newblock In \emph{ICCV}, pages 21525--21535, 2025.

\bibitem[{NVIDIA}(2026)]{nvidia2026ultrasoundsimulator}
{NVIDIA}.
\newblock {NVIDIA Isaac for Healthcare}: Raytracing ultrasound simulator.
\newblock Software documentation, 2026.
\newblock Accessed: 2026-11-15.

\bibitem[{NVIDIA} et~al.(2025){NVIDIA}, Agarwal, Ali, Bala, Balaji,
  et~al.]{nvidia2025cosmos}
{NVIDIA}, Niket Agarwal, Arslan Ali, Maciej Bala, Yogesh Balaji, et~al.
\newblock {Cosmos} world foundation model platform for physical {AI}.
\newblock \emph{arXiv preprint arXiv:2501.03575}, 2025.

\bibitem[Phi et~al.(2024)Phi, Duc, Hieu, and Long]{phi2024echosynthesis}
Nguyen~Van Phi, Tran~Minh Duc, Pham~Huy Hieu, and Tran~Quoc Long.
\newblock Echocardiography video synthesis from end diastolic semantic map via
  diffusion model.
\newblock In \emph{IEEE International Conference on Acoustics, Speech and
  Signal Processing}, 2024.

\bibitem[Salehi et~al.(2015)Salehi, Ahmadi, Prevost, Navab, and
  Wein]{salehi2015simulation}
Mehrdad Salehi, Seyed-Ahmad Ahmadi, Raphael Prevost, Nassir Navab, and Wolfgang
  Wein.
\newblock Patient-specific 3d ultrasound simulation based on convolutional
  ray-tracing and appearance optimization.
\newblock In \emph{Medical Image Computing and Computer-Assisted Intervention},
  pages 510--518, 2015.

\bibitem[Sim{\'e}oni et~al.(2025)Sim{\'e}oni, Vo, Seitzer, Baldassarre, Oquab,
  Jose, Khalidov, Szafraniec, Yi, Ramamonjisoa, et~al.]{simeoni2025dinov3}
Oriane Sim{\'e}oni, Huy~V. Vo, Maximilian Seitzer, Federico Baldassarre, Maxime
  Oquab, Cijo Jose, Vasil Khalidov, Marc Szafraniec, Seungeun Yi, Micha{\"e}l
  Ramamonjisoa, et~al.
\newblock {DINOv3}.
\newblock \emph{arXiv preprint arXiv:2508.10104}, 2025.

\bibitem[Su et~al.(2024)Su, Ahmed, Lu, Pan, Bo, and Liu]{su2024rope}
Jianlin Su, Murtadha Ahmed, Yu Lu, Shengfeng Pan, Wen Bo, and Yunfeng Liu.
\newblock {RoFormer}: Enhanced transformer with rotary position embedding.
\newblock \emph{Neurocomputing}, 568, 2024.

\bibitem[{Team Wan} et~al.(2025)]{wan2025}
{Team Wan} et~al.
\newblock Wan: Open and advanced large-scale video generative models.
\newblock \emph{arXiv preprint arXiv:2503.20314}, 2025.

\bibitem[Vitale et~al.(2020)Vitale, Orlando, Iarussi, and
  Larrabide]{vitale2020simulation}
Santiago Vitale, Jos{\'e}~Ignacio Orlando, Emmanuel Iarussi, and Ignacio
  Larrabide.
\newblock Improving realism in patient-specific abdominal ultrasound simulation
  using {CycleGANs}.
\newblock \emph{International Journal of Computer Assisted Radiology and
  Surgery}, 15\penalty0 (2):\penalty0 183--192, 2020.

\bibitem[Wasserthal et~al.(2023)Wasserthal, Breit, Meyer, Pradella, Hinck,
  Sauter, Heye, Boll, Cyriac, Yang, et~al.]{wasserthal2023totalsegmentator}
Jakob Wasserthal, Hanns-Christian Breit, Manfred~T. Meyer, Maurice Pradella,
  Daniel Hinck, Alexander~W. Sauter, Tobias Heye, Daniel~T. Boll, Joshy Cyriac,
  Shan Yang, et~al.
\newblock {TotalSegmentator}: Robust segmentation of 104 anatomic structures in
  {CT} images.
\newblock \emph{Radiology: Artificial Intelligence}, 5\penalty0 (5), 2023.

\bibitem[Williams and Zipser(1989)]{williams1989teacherforcing}
Ronald~J. Williams and David Zipser.
\newblock A learning algorithm for continually running fully recurrent neural
  networks.
\newblock \emph{Neural Computation}, 1\penalty0 (2):\penalty0 270--280, 1989.

\bibitem[Xiao et~al.(2024)Xiao, Tian, Chen, Han, and Lewis]{xiao2024sink}
Guangxuan Xiao, Yuandong Tian, Beidi Chen, Song Han, and Mike Lewis.
\newblock Efficient streaming language models with attention sinks.
\newblock In \emph{ICLR}, 2024.

\bibitem[Xu et~al.(2025)Xu, Dong, Yang, et~al.]{xu2025vsnet}
Jichen Xu, Anqi Dong, Yang Yang, et~al.
\newblock {VSNet}: Vessel structure-aware network for hepatic and portal vein
  segmentation.
\newblock \emph{Medical Image Analysis}, 101:\penalty0 103458, 2025.

\bibitem[Yang et~al.(2026{\natexlab{a}})Yang, Wang, Guan, and
  Ren]{yang2026ultraworld}
Keke Yang, Erqi Wang, Sainan Guan, and Hongliang Ren.
\newblock Ultraworld: Learning interactive ultrasound world models from
  untracked clinical videos with acoustic sampling map.
\newblock \emph{arXiv preprint arXiv:2610.09785}, 2026{\natexlab{a}}.

\bibitem[Yang et~al.(2026{\natexlab{b}})Yang, Zhang, Maihe, Liu, Yu, Mukherjee,
  Chen, Yan, Chen, and Sun]{yang2026sonoworld}
Zefan Yang, Zheyuan Zhang, Boris Maihe, Yikang Liu, Hang Yu, Ankush Mukherjee,
  Xiao Chen, Pingkun Yan, Terrence Chen, and Shanhui Sun.
\newblock Sonoworld: Towards interactive pose-controllable ultrasound world
  models.
\newblock \emph{arXiv preprint arXiv:2610.19327}, 2026{\natexlab{b}}.

\bibitem[Yin et~al.(2024)Yin, Gharbi, Zhang, Shechtman, Durand, Freeman, and
  Park]{yin2024dmd}
Tianwei Yin, Micha{\"e}l Gharbi, Richard Zhang, Eli Shechtman, Fr{\'e}do
  Durand, William~T. Freeman, and Taesung Park.
\newblock One-step diffusion with distribution matching distillation.
\newblock In \emph{CVPR}, 2024.

\bibitem[Yin et~al.(2025)Yin, Zhang, Zhang, Freeman, Durand, Shechtman, and
  Huang]{yin2025causvid}
Tianwei Yin, Qiang Zhang, Richard Zhang, William~T. Freeman, Fr{\'e}do Durand,
  Eli Shechtman, and Xun Huang.
\newblock From slow bidirectional to fast autoregressive video diffusion
  models.
\newblock In \emph{CVPR}, 2025.

\bibitem[Yue et~al.(2025{\natexlab{a}})Yue, Wang, Jiang, Liu, Song, and
  Huang]{yue2025echoworld}
Yang Yue, Yulin Wang, Haojun Jiang, Pan Liu, Shiji Song, and Gao Huang.
\newblock Echoworld: Learning motion-aware world models for echocardiography
  probe guidance.
\newblock In \emph{CVPR}, pages 25993--26003, 2025{\natexlab{a}}.

\bibitem[Yue et~al.(2025{\natexlab{b}})Yue, Wang, Tao, Liu, Song, and
  Huang]{yue2025chexworld}
Yang Yue, Yulin Wang, Chenxin Tao, Pan Liu, Shiji Song, and Gao Huang.
\newblock {CheXWorld}: Exploring image world modeling for radiograph
  representation learning.
\newblock In \emph{CVPR}, 2025{\natexlab{b}}.

\bibitem[Zhang et~al.(2023)Zhang, Rao, and Agrawala]{zhang2023controlnet}
Lvmin Zhang, Anyi Rao, and Maneesh Agrawala.
\newblock Adding conditional control to text-to-image diffusion models.
\newblock In \emph{ICCV}, 2023.

\bibitem[Zhang et~al.(2018)Zhang, Isola, Efros, Shechtman, and
  Wang]{zhang2018lpips}
Richard Zhang, Phillip Isola, Alexei~A. Efros, Eli Shechtman, and Oliver Wang.
\newblock The unreasonable effectiveness of deep features as a perceptual
  metric.
\newblock In \emph{CVPR}, 2018.

\bibitem[Zhang et~al.(2025)Zhang, Yao, Wang, Jha, Durak, Keles,
  Medetalibeyoglu, and Bagci]{zhang2025diffboost}
Zheyuan Zhang, Lanhong Yao, Bin Wang, Debesh Jha, Gorkem Durak, Elif Keles,
  Alpay Medetalibeyoglu, and Ulas Bagci.
\newblock {DiffBoost}: Enhancing medical image segmentation via text-guided
  diffusion model.
\newblock \emph{IEEE Transactions on Medical Imaging}, 44\penalty0
  (9):\penalty0 3670--3682, 2025.

\end{thebibliography}

}
\end{document}